\documentclass[11pt]{article}

\usepackage[final]{acl}

\usepackage{times}
\usepackage{latexsym}

\usepackage[T1]{fontenc}

\usepackage[utf8]{inputenc}

\usepackage{microtype}

\usepackage{inconsolata}

\usepackage[most]{tcolorbox}

\usepackage{hyperref}
\usepackage{url}
\usepackage{wrapfig}
\usepackage{algorithm}
\usepackage{booktabs} 
\usepackage{tabularx} 
\usepackage{multirow}
\usepackage{algpseudocode}
\usepackage{float}
\usepackage{setspace}
\usepackage{threeparttable}
\usepackage[normalem]{ulem}
\usepackage[table]{xcolor}
\usepackage{arydshln, colortbl}

\usepackage{xcolor}
\usepackage{colortbl}
\usepackage{soul}

\usepackage{tikz}
\usetikzlibrary{positioning,calc}
\usetikzlibrary{mindmap}

\tcbuselibrary{breakable}

\definecolor{SDEblue}{RGB}{28 58 88}
\definecolor{lighterblue}{HTML}{f2fafd}
\definecolor{cc3}{RGB}{255, 191, 0}
\definecolor{cc4}{RGB}{0, 128, 128}
\hypersetup{
 colorlinks=true,
 linkcolor=SDEblue,
 urlcolor=magenta,
 citecolor=SDEblue,
}
\newcommand{\Seclink}[1]{\hyperref[#1]{\S~\ref*{#1}}}

\definecolor{customorange}{HTML}{ffe6d5}

\definecolor{custompink}{HTML}{ffd5d5}

\usepackage{adjustbox}
\usepackage{tikz}
\usepackage[edges]{forest}
\definecolor{hidden-draw}{RGB}{64,101,149}
\definecolor{hidden-pink}{RGB}{231,239,250}
\definecolor{darkgray}{RGB}{60, 60, 60}

\usepackage{pifont}
\definecolor{myblue}{HTML}{AFC6E9}

\newcommand{\blueyes}{\textcolor{myblue}{\ding{51}}}

\usepackage{twemojis}

\definecolor{Blueblock}{HTML}{b3c5db}
\definecolor{Orangeblock}{HTML}{dabda5}
\DeclareRobustCommand{\ColorBox}[1]{%
  \tikz[baseline=0ex, inner sep=0pt, outer sep=0pt]%
  \filldraw[fill=#1, draw=black, line width=0.6pt, rounded corners=0.4ex] (0,0) rectangle (2.0ex,2.0ex);%
}
\DeclareRobustCommand{\BlueBlock}{\ColorBox{Blueblock}}
\DeclareRobustCommand{\OrangeBlock}{\ColorBox{Orangeblock}}
\DeclareRobustCommand{\SolidArrow}{%
  \tikz[baseline=-0.6ex, inner sep=0pt, outer sep=0pt]%
  \draw[->, >=stealth, line width=0.8pt, black] (0,0) -- (2.5em, 0);%
}

\DeclareRobustCommand{\DashedArrow}{%
  \tikz[baseline=-0.6ex, inner sep=0pt, outer sep=0pt]%
  \draw[->, >=stealth, dashed, dash pattern=on 3pt off 2pt, line width=0.8pt, black] (0,0) -- (2.5em, 0);%
}

\title{
    \raisebox{-0.15em}{\includegraphics[height=1.2em]{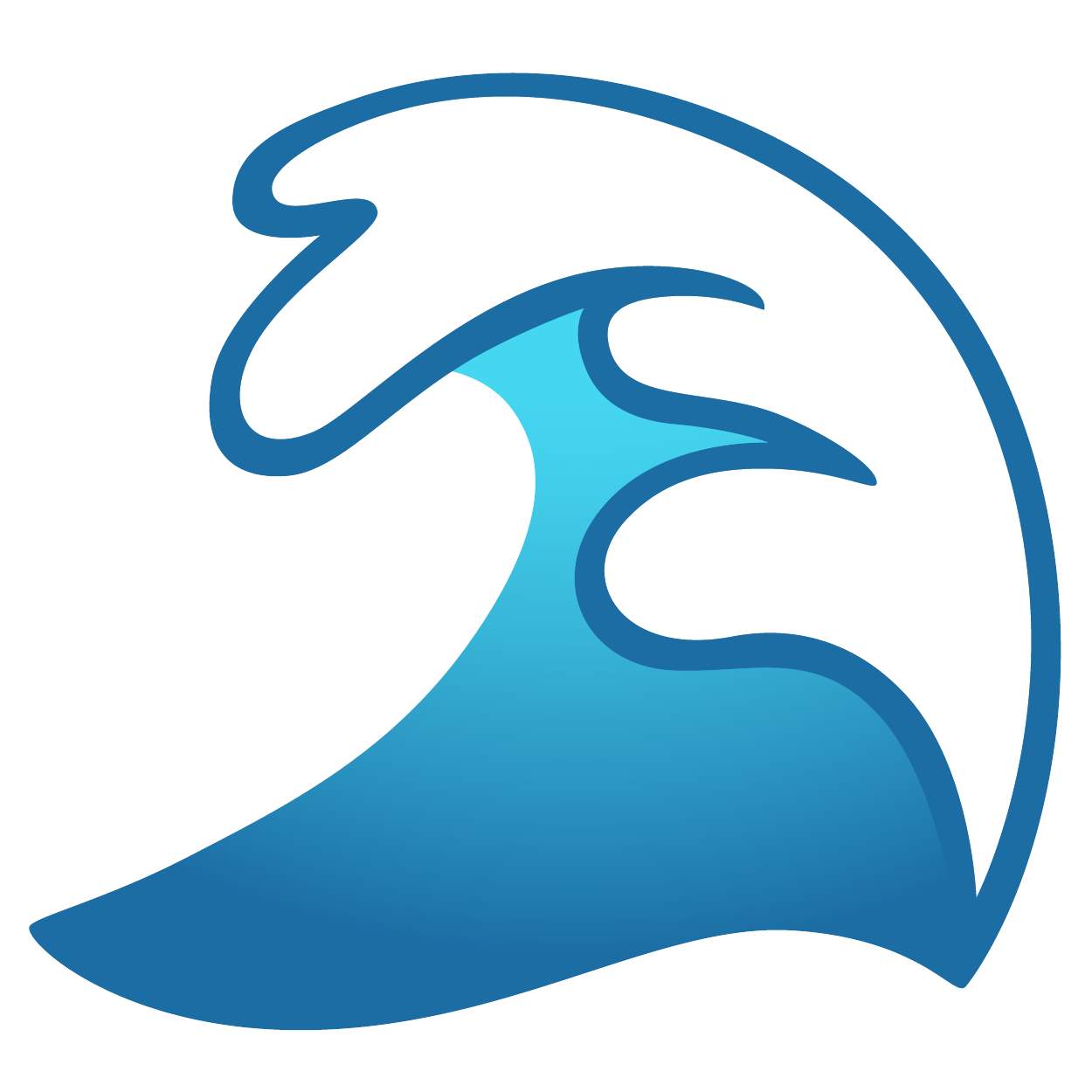}}
  From Static Inference to Dynamic Interaction:\\ A Survey of Streaming Large Language Models
}

\author{
 \textbf{Junlong Tong\textsuperscript{1,2}},
 \textbf{Zilong Wang\textsuperscript{2}},
 \textbf{YuJie Ren\textsuperscript{2}},
 \textbf{Peiran Yin\textsuperscript{2}},\\
 \textbf{Hao Wu\textsuperscript{2}},
 \textbf{Wei Zhang\textsuperscript{2}},
 \textbf{Xiaoyu Shen\textsuperscript{2}\thanks{Corresponding author}}
\\
 \textsuperscript{1}Shanghai Jiao Tong University\\
 \textsuperscript{2}Institute of Digital Twin, Eastern Institute of Technology, Ningbo
\\
\texttt{jl-tong@sjtu.edu.cn}~~~~~\texttt{xyshen@eitech.edu.cn}
}

\begin{document}
\maketitle

\begin{abstract}

Standard Large Language Models (LLMs) are predominantly designed for static inference with pre-defined inputs, which limits their applicability in dynamic, real-time scenarios. To address this gap, the \emph{streaming LLM} paradigm has emerged.
However, existing definitions of streaming LLMs remain fragmented, conflating streaming generation, streaming inputs, and interactive streaming architectures, while a systematic taxonomy is still lacking.
This paper provides a comprehensive overview and analysis of streaming LLMs. 
First, we establish a unified definition of streaming LLMs based on data flow and dynamic interaction to clarify existing ambiguities.
Building on this definition, we propose a systematic taxonomy of current streaming LLMs and provide an in-depth discussion of their underlying methodologies across text, speech, and video streaming scenarios.
Furthermore, we explore the applications of streaming LLMs in real-world scenarios and outline promising research directions to support ongoing advances in streaming intelligence.
We maintain a continuously updated repository of relevant papers at \url{https://github.com/EIT-NLP/Awesome-Streaming-LLMs}.

\end{abstract}
\section{Introduction}

Large Language Models (LLMs) have shown remarkable efficacy across diverse domains, exhibiting strong reasoning, generation, and cross-modal capabilities~\cite{OpenAI2023,team2023gemini,deepseekv3}. However, LLMs are predominantly pre-trained on \emph{static and full-context corpora}, following a ``read-at-once'' paradigm in which the complete input is provided before any output is generated. While effective for benchmark-style tasks, this paradigm fundamentally limits their applicability in real-world environments, where information arrives incrementally, accumulates over time, and may be unbounded in length.

Such dynamic conditions are ubiquitous in tasks like real-time translation, streaming video understanding, and interactive tool agents~\cite{agostinelli2024simul,jin2025streamingassistant,yang2025streamagent}. In these real-world applications, inputs such as speech and video data stream continuously, forcing systems to maintain an evolving understanding based on partial observations.
In more complex scenarios, these signals may originate from \emph{multiple concurrent streams}~\cite{li2025watch}, while systems may also need to generate multiple outputs in parallel~\cite{zhang2025stream}. For instance, a robot may need to act, speak, and reason simultaneously~\cite{zhang2025activevln}, whereas an interactive assistant may coordinate speech, visual updates, and control commands~\cite{zhang2025avila}. Since the input is never fully available at any given moment, the system must dynamically decide \emph{when to respond, when to wait for more information, and when to terminate}~\cite{panchal2024say,zhang2025proactive}. These requirements expose a fundamental mismatch with the offline, full-context design of standard LLMs.

\begin{figure}[t]
\begin{center}
\includegraphics[width=1\linewidth]{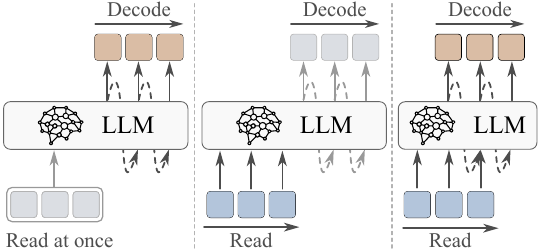}
\end{center}
\caption{\small Illustration of three types of streaming large language models (LLMs). (\textit{Left}) \textbf{Output-streaming LLM} performs streaming generation \textit{after} static reading. (\textit{Middle}) \textbf{Sequential-streaming LLM} performs streaming generation \textit{after} streaming reading. (\textit{Right}) \textbf{Concurrent-streaming LLM} performs streaming generation \textit{while} streaming reading.}
\label{Fig:Top}
\end{figure}

Adapting LLMs to these real-world streaming scenarios presents significant challenges.
Beyond architectural modifications, there is a scarcity of large-scale pre-training data that supports real-time interaction, partial-input supervision, and fine-grained temporal alignment. Motivated by this gap, recent research has begun to investigate \emph{streaming LLMs}~\cite{tong2025streamingthinker,chen2024videollm,du2024cosyvoice}. However, the field currently suffers from terminological ambiguity. Existing studies often conflate distinct concepts, such as \emph{autoregressive decoding}~\cite{kondratyuk2024videopoet}, \emph{incremental or chunk-wise encoding}~\cite{xiao2023streamingllm}, and \emph{full-duplex interaction} like GPT-4o~\cite{OpenAI2023}, under a single ``Streaming LLM'' umbrella, obscuring meaningful comparisons.

In this work, we provide the first systematic review of streaming LLMs, proposing a unified definition based on data flow and interaction concurrency. As illustrated in~\autoref{Fig:Top}, we categorize these models into three distinct levels:
(1) \textit{Output-streaming LLMs}, which retain static input processing but support streaming output generation.
(2) \textit{Sequential-streaming LLMs}, which process streaming inputs incrementally but generate with full input.
(3) \textit{Concurrent-streaming LLMs}, which enable full-duplex interaction by continuously receiving inputs and generating outputs.

This taxonomy captures both conceptual distinctions and a clear progression of technical challenges: Output streaming addresses challenges in streaming and low-latency generation; sequential streaming introduces incremental encoding and context management; and concurrent streaming builds upon both to address architecture adaptation and interaction strategies required for full-duplex processing. By disentangling these paradigms, the taxonomy clarifies which challenges are shared, which are incremental, and which are unique to each category, thereby providing a structured roadmap toward the ultimate goal of fully interactive streaming LLMs.
Guided by this framework, we systematically review representative methods in each category, examine emerging applications such as streaming video understanding and real-time reasoning, and highlight open problems, including trade-offs between latency and performance, to inform future research.\footnote{We provide a detailed description of motivation, survey scope, and difference with related surveys in Appendix~\ref{Appendix-A}.}

To summarize, our main contributions include: 
\vspace{-0.5em}
\begin{itemize}
    \setlength{\itemsep}{1pt}
    \setlength{\parskip}{1pt}
    \setlength{\parsep}{1pt}

    \item To our knowledge, we are the \textbf{first systematic survey} of streaming LLMs.
    \item We introduce \textbf{a unified definition} of streaming LLMs, clarifying the conceptual distinctions among existing paradigms.
    \item We provide \textbf{a systematic taxonomy and comprehensive technical analysis}, disentangling the mechanisms of three streaming paradigms.
    \item We discuss \textbf{emerging applications and open research directions} for real-time and interactive streaming scenarios.
\end{itemize}

\begin{figure*}[t]
\begin{center}
\includegraphics[width=\linewidth]{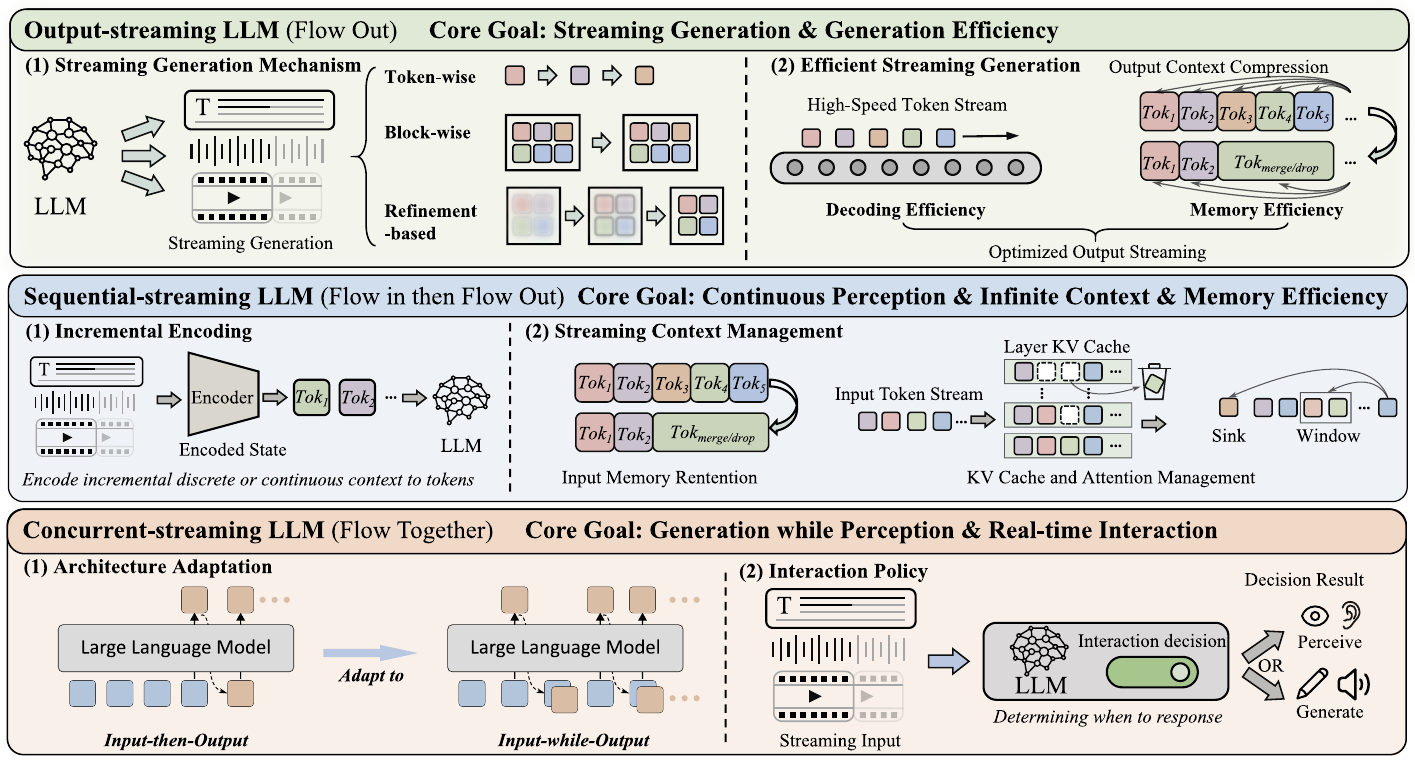}
\end{center}
\caption{Overview of streaming LLM paradigms and their key challenges. The figure contrasts Output-streaming, Sequential-streaming, and Concurrent-streaming LLMs, highlighting their core goals and corresponding research components. Concurrent-streaming builds on the first two and adds extra challenges in real-time streaming architecture adaptation and interaction policy learning.}
\label{Fig:Main}
\end{figure*}
\section{Preliminaries}
\subsection{Background of Streaming LLMs}
Current LLMs typically operate under a batch processing paradigm, where the model encodes the entire input sequence into the KV cache during the prefill phase and subsequently generates tokens autoregressively in the decoding phase. Consequently, from a data flow perspective, standard LLMs can be categorized as ``streaming-output LLMs'' that rely on static context availability. However, real-world data flows often exhibit dynamic and continuous characteristics (e.g., real-time speech transcription and content understanding), necessitating models capable of handling streaming inputs and executing timely output decisions; therefore, generalized streaming LLMs are defined to address such dynamic input and immediate response scenarios, aiming to transcend the limitations of static preprocessing and delayed response.

\tikzstyle{my-box}=[
    rectangle,
    draw=hidden-draw,
    rounded corners,
    text opacity=1,
    minimum height=1.5em,
    minimum width=5em,
    inner sep=2pt,
    align=center,
    fill opacity=.5,
    line width=0.8pt,
]
\tikzstyle{leaf}=[my-box, minimum height=1.5em,
    fill=hidden-pink!80, text=black, align=left,font=\normalsize,
    inner xsep=2pt,
    inner ysep=4pt,
    line width=0.8pt,
]

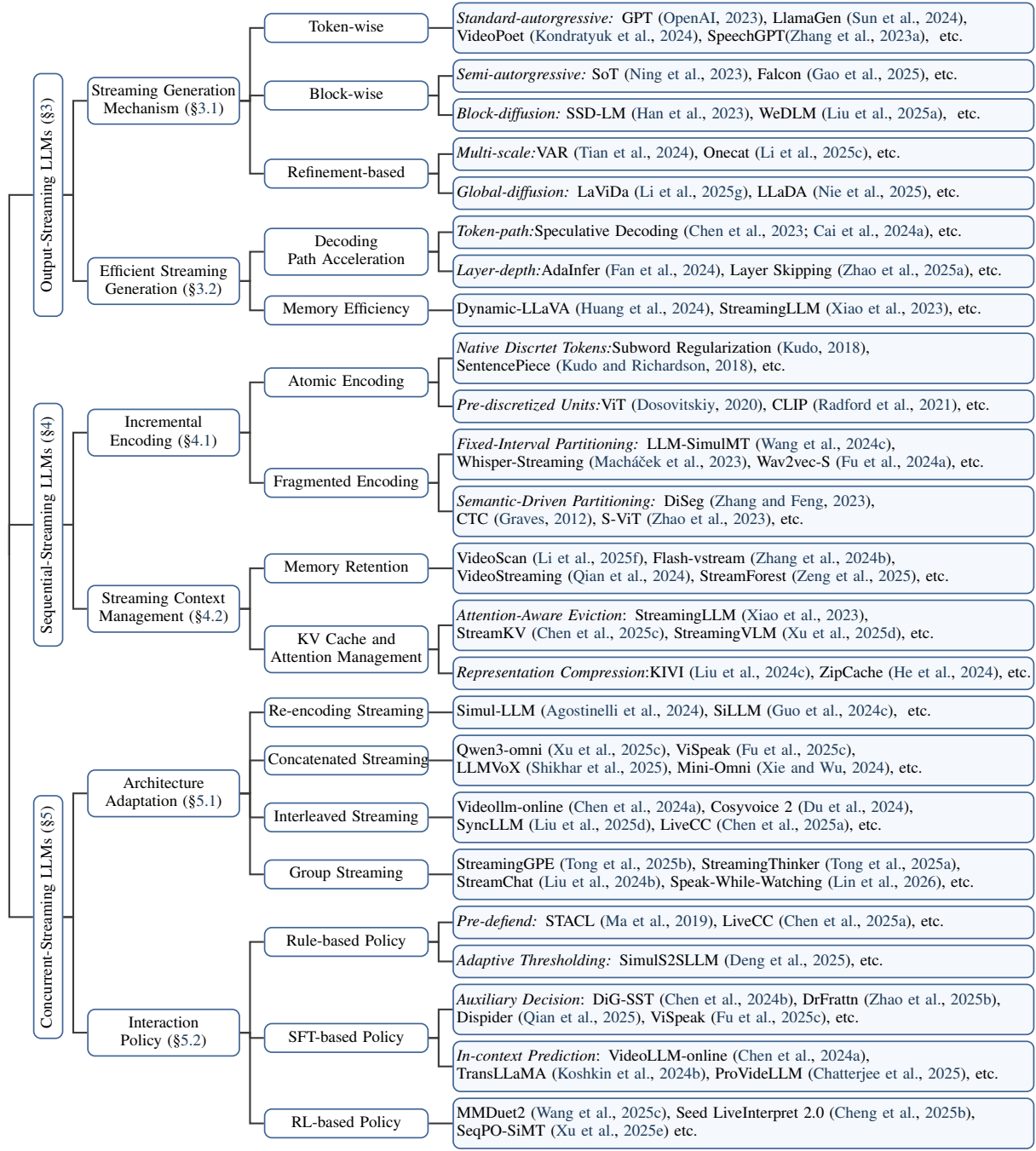
\begin{figure*}[t!]
    \centering
    \resizebox{0.9\textwidth}{!}{
        \begin{forest}
            forked edges,
            for tree={
                grow=east,
                reversed=true,
                anchor=base west,
                parent anchor=east,
                child anchor=west,
                base=center,
                font=\large,
                rectangle,
                draw=hidden-draw,
                rounded corners,
                align=left,
                text centered,
                minimum width=0.2em,
                edge+={darkgray, line width=1pt},
                s sep=3pt,
                inner xsep=2pt,
                inner ysep=3pt,
                line width=0.8pt,
                ver/.style={rotate=90, child anchor=north, parent anchor=south, anchor=center},
            },
            where level=1{text width=12em,font=\small}{},
            where level=2{text width=7.3em,font=\small}{},
            where level=3{text width=7.9em,font=\small}{},
            where level=4{text width=8em,font=\small}{},
            [
                , coordinate, for children={fork sep=0pt, l=0.1em}
                [
                    Output-Streaming LLMs (\S\ref{Sec:output-streaming}), ver, align=center,for children={fork sep=0.5em, l=1em}
                    [
                        Streaming Generation\\ Mechanism (\S \ref{subsec:streaming_generation_mechanism}), align=center
                        [
                            Token-wise
                            [   
                                \textit{Standard-autorgressive:} GPT~\cite{OpenAI2023}{, }LlamaGen~\cite{sun2024autoregressive}{, }\\VideoPoet~\cite{kondratyuk2024videopoet}{, }SpeechGPT\cite{zhang2023speechgpt}{, } etc.
                                , leaf, text width=29em, font=\small
                            ]
                        ]
                        [
                            Block-wise\\
                            [
                                \textit{Semi-autorgressive:}~SoT~\cite{ning2023skeleton}{, }Falcon~\cite{gao2025falcon}{,} etc.\\
                                , leaf, text width=29em, font=\small
                            ]
                            [
                                \textit{Block-diffusion:}~SSD-LM~\cite{han2023ssd}{, }WeDLM~\cite{liu2025WeDLM}{, } etc.\\
                                , leaf, text width=29em, font=\small
                            ]                            
                        ]                        
                        [
                            Refinement-based\\
                            [
                                \textit{Multi-scale:}VAR~\cite{tian2024visual}{, }Onecat~\cite{li2025onecat}{, }etc.
                                , leaf, text width=29em, font=\small
                            ]                            
                            [
                                \textit{Global-diffusion:} LaViDa~\cite{li2025lavida}{, }LLaDA~\cite{nie2025large}{, }etc.
                                , leaf, text width=29em, font=\small
                            ]
                        ]
                    ]
                    [
                        Efficient Streaming\\ Generation (\S \ref{subsec:efficient_streaming_generation}), align=center
                        [    Decoding \\Path Acceleration, align=center
                            [
                                \textit{Token-path:}Speculative Decoding~\cite{chen2023accelerating,cai2024medusa}{, }etc.
                                , leaf, text width=29em, font=\small
                            ]
                            [
                                \textit{Layer-depth:}AdaInfer~\cite{fan2024not}{, }Layer Skipping~\cite{zhao2025skipgpt}{, }etc.
                                , leaf, text width=29em, font=\small
                            ]
                        ]               
                        [
                            Memory Efficiency\\
                            [
                                Dynamic-LLaVA~\cite{huang2024dynamic}{, }StreamingLLM~\cite{xiao2023streamingllm}{, }etc.
                                , leaf, text width=29em, font=\small
                            ]
                        ]
                    ]
                ]
                [
                    Sequential-Streaming LLMs (\S \ref{Sec:sequential-streaming}), ver, align=center,for children={fork sep=0.5em, l=1em}
                    [
                        Incremental\\ Encoding (\S \ref{subsec:incremental-encoding}), align=center
                        [
                            Atomic Encoding
                            [
                                \textit{Native Discrtet Tokens:}Subword Regularization~\citep{kudo2018subword}{,} \\SentencePiece~\citep{kudo2018sentencepiece}{,} etc.
                                , leaf, text width=29em, font=\small
                            ]
                            [
                                \textit{Pre-discretized Units:}ViT~\cite{dosovitskiy2020image}{,} CLIP~\cite{radford2021learning}{,} etc.
                                , leaf, text width=29em, font=\small
                            ]
                        ]
                        [
                            Fragmented Encoding\\
                            [
                                \textit{Fixed-Interval Partitioning:} LLM-SimulMT~\cite{wang2024simultaneous}{,} \\Whisper-Streaming~\cite{machavcek2023turning}{, }Wav2vec-S~\cite{fu2024wav2vec}{, }etc.
                                , leaf, text width=29em, font=\small
                            ]
                           [
                                \textit{Semantic-Driven Partitioning:} DiSeg~\cite{zhang2023end}{,}\\CTC~\cite{graves2012connectionist}{,} S-ViT~\cite{zhao2023streaming}{,} etc.
                                , leaf, text width=29em, font=\small
                            ]
                        ]
                    ]
                    [
                        Streaming Context\\ Management (\S \ref{subsec:context-management}), align=center
                        [
                            Memory Retention, align=center
                            [
                                VideoScan~\cite{li2025videoscan}{,} Flash-vstream~\cite{zhang2024flash}{,} \\ VideoStreaming~\cite{qian2024streaming}{,} StreamForest~\cite{zeng2025streamforest}{, }etc.
                                , leaf, text width=29em, font=\small
                            ]
                        ]
                        [
                            KV Cache and\\Attention Management, align=center
                            [
                               \textit{Attention-Aware Eviction}: StreamingLLM~\cite{xiao2023streamingllm}{, }\\StreamKV~\cite{chen2025streamkv}{, }StreamingVLM~\cite{xu2025streamingvlm}{,} etc.
                                , leaf, text width=29em, font=\small
                            ]
                            [
                                \textit{Representation Compression}:KIVI~\cite{liu2024kivi}{,} ZipCache~\cite{he2024zipcache}{,} etc.
                                , leaf, text width=29em, font=\small
                            ]
                        ]
                    ]
                ]
                [
                    Concurrent-Streaming LLMs (\S \ref{Sec:Concurrent-streaming}), ver, align=center,for children={fork sep=0.5em, l=1em}
                    [
                        Architecture\\ Adaptation (\S \ref{subsec:architecutre_adaptation}), align=center
                        [
                           Re-encoding Streaming\\, align=center
                            [
                                Simul-LLM~\cite{agostinelli2024simul}{, }SiLLM~\cite{guo2024sillm}{, } etc.
                                , leaf, text width=29em, font=\small
                            ]
                        ]
                        [
                           Concatenated Streaming\\, align=center
                            [
                                Qwen3-omni~\cite{Qwen3-Omni}{, }ViSpeak~\cite{fu2025vispeak}{, }\\LLMVoX~\cite{shikhar2025llmvox}{, }Mini-Omni~\cite{xie2024mini}{, }etc.
                                , leaf, text width=29em, font=\small
                            ]
                        ]
                        [
                           Interleaved Streaming\\, align=center
                            [
                                Videollm-online~\cite{chen2024videollm}{, }Cosyvoice 2~\cite{du2024cosyvoice}{, }\\SyncLLM~\cite{liu2025sync}{, }LiveCC~\cite{chen2025livecc}{, }etc.
                                , leaf, text width=29em, font=\small
                            ]
                        ]
                        [
                            Group Streaming\\, align=center
                            [
                                StreamingGPE~\cite{tong2025llm}{, }StreamingThinker~\cite{tong2025streamingthinker}{, }\\StreamChat~\cite{liu2024streamchat}{, }Speak-While-Watching~\cite{lin2026speak}{, }etc.
                                , leaf, text width=29em, font=\small
                            ]
                        ]
                    ]
                    [
                        Interaction\\ Policy (\S \ref{subsec:interaction_policy}), align=center
                        [
                            Rule-based Policy\\, align=center
                            [
                                \textit{Pre-defiend:} STACL~\cite{ma2019stacl}{, }LiveCC~\cite{chen2025livecc}{, }etc.\\
                                , leaf, text width=29em, font=\small
                            ]
                            [
                                \textit{Adaptive Thresholding:} SimulS2SLLM~\cite{deng2025simuls2s}{, }etc.
                                , leaf, text width=29em, font=\small
                            ]
                        ]
                        [
                            SFT-based Policy\\, align=center
                            [
                                \textit{Auxiliary Decision}: DiG-SST~\cite{chen2024divergence}{, }DrFrattn~\cite{zhao2025drfrattn}{,}\\
                                Dispider~\cite{qian2025dispider}{, }ViSpeak~\cite{fu2025vispeak}{, }etc.
                                , leaf, text width=29em, font=\small
                            ]
                            [
                                \textit{In-context Prediction}: VideoLLM-online~\cite{chen2024videollm}{, }\\TransLLaMA~\cite{koshkin2024transllama}{, }ProVideLLM~\cite{chatterjee2025streaming}{, }etc.
                                , leaf, text width=29em, font=\small
                            ]
                        ]
                        [
                            RL-based Policy\\, align=center
                            [
                                MMDuet2~\cite{wang2025mmduet2}{, }Seed LiveInterpret 2.0~\cite{cheng2025seed}{,}\\
                                SeqPO-SiMT~\cite{xu2025seqpo} etc.
                                , leaf, text width=29em, font=\small
                            ]
                        ]
                    ]
                ]
            ]
        \end{forest}
    }
    \caption{\small Taxonomy of Streaming Large Language Models.}
    \label{fig:taxo_of_streaming_llms}
\end{figure*}

\subsection{Formal Definition}
To rigorously unify the diverse landscape of streaming LLMs in~\autoref{Fig:Top}, we formulate the modeling process as a conditional probability distribution $P(Y|X)$, where $X = (x_1, \dots, x_M)$ denotes the bounded input stream and $Y = (y_1, \dots, y_N)$ denotes the output stream.
This distribution can be factorized autoregressively using the chain rule:
\begin{equation}
    P(Y|X) = \prod_{t=1}^{N} P\big(y_t | y_{<t}, h_{1:\phi(t)}(X);\theta\big)
\end{equation}
where $\theta$ denotes the LLM parameters, and $h_{\phi(t)}(X)=llm(x_{\phi(t)})$ represents the encoded hidden states corresponding to the input prefix $x_{\phi(t)}$. 
Here, $\phi(t)$ is a decision function to determine the input stream visible at generation step $t$.
This general definition can be instantiated into three sub-types by applying varying operational constraints.

\paragraph{Output-streaming LLMs}
This paradigm imposes a static constraint where the entire input must be processed before generation begins. Mathematically, the decision function is constant relative to the total input length $M$, i.e., $\phi(t) = M$ for all $t \in \{1, \dots, N\}$. The hidden states are computed via a one-time global prefilling: $h_{1:\phi(t)}(X) = h_{1:M}(X) = llm(X_{1:M}).$

\paragraph{Sequential-streaming LLMs}
This paradigm processes dynamic streaming inputs but generates based on a fixed input. While the decision function mirrors the above type (i.e., $\phi(t) = M, \forall t$), the hidden states are constrained by stepwise arrival: $h_{1:M}(X) = \{ llm(x_1), \dots, llm(x_M) \}.$ This represents a sequential encoding process where the context is accumulated token-by-token (or chunk-by-chunk) before the generation phase begins.

\paragraph{Concurrent-streaming LLMs}
This paradigm imposes the strictest temporal constraints, representing a dynamic process where streams unfold continuously. Mathematically, $\phi(t)$ must satisfy monotonicity and partial visibility:$1\le \dots \le \phi(t)\le \phi(t+1)\le \dots \le M.$
The hidden states of input stream are computed via a dynamic or interactive process: $h_{\phi(t)}(X) = llm(X_{\phi(t)},y_{<t}).$

The tripartite taxonomy defined above reflects a trajectory of escalating operational constraints and functional demands, shifting the paradigm from static processing to dynamic, real-time interaction. 

$1\le \dots \le \phi(t)\le \phi(t+1)\le \dots \le M.$

\subsection{Overview}
This survey provides a systematical overview of research in streaming LLMs.
Figure~\ref{Fig:Main} illustrates the proposed taxonomy, detailing the primary research focuses and challenges within each category. Specifically, output-streaming emphasizes streaming generation mechanisms and efficient generation; sequential-streaming focuses on incremental encoding processing and context management for input streams; and concurrent-streaming integrates both tasks, additionally introducing architectural adaptations and the interactive management of simultaneous input and output streams.
To navigate this comprehensive landscape, \autoref{fig:taxo_of_streaming_llms} outlines the taxonomy structure of this survey. Guided by this taxonomy, we begin with output-streaming in Section 3, expand to the dynamic input processing of sequential-streaming in Section 4, and culminate with the interactive dynamics of concurrent-streaming in Section 5. Beyond the technical part, Section 6 reviews downstream tasks and applications, and Section 7 discusses the future directions.

\section{Output-Streaming LLMs: Generating with Progressive Revelation}
\label{Sec:output-streaming}
\subsection{Streaming Generation Mechanism}
\label{subsec:streaming_generation_mechanism}

Output streaming enables \textit{progressive revelation} by continuously emitting intermediate results rather than waiting for completion. Based on the generation granularity and update mechanism, we categorize existing methods into: (i) \textbf{token-wise}, (ii) \textbf{block-wise}, and (iii) \textbf{refinement-based}.

\paragraph{Token-wise}
This represents the dominant generation paradigm for LLMs, employing token-wise autoregressive decoding~\cite{team2023gemini,deepseekv3,gemma}.
For multimodal outputs, systems typically extend this paradigm by aligning non-text modalities to the textual space for autoregressive streaming~\cite{zhang2023speechgpt,llamagen2024}.

\paragraph{Block-wise}
These methods expand the generation unit from single tokens to multi-token blocks, reducing serial depth while retaining the controllability of autoregressive modeling. We summarized them into two lines.
(1) \textit{Semi-autoregressive} relaxes intra-block dependencies to predict multiple tokens \textit{in parallel}.~\cite{wang2018sat, hwang2025predsent,ning2023skeleton,gao2025falcon}.
For example, MTP~\cite{gloeckle2024better} predicts multiple tokens simultaneously for each autoregressive block step.
(2) \textit{Block-diffusion} combines diffusion-style refinement with block-wise generation, iteratively \textit{denoising a block} at a time and streaming blocks autoregressively~\cite{han2023ssd,liu2025WeDLM,tian2025next,arriola2025block}.

\paragraph{Refinement-based}
Unlike token-by-token sequential accumulation, this paradigm performs progressive refinement \textit{from coarse to fine}, iteratively improving the semantic completeness of the entire sequence rather than merely extending its length.
(1) \textit{Multi-scale} approach decomposes generation into discrete scales~\cite{tian2024visual,li2025onecat,zhuang2025vargpt}. Models like VAR~\cite{tian2024visual} predict the next-scale autoregressively, enabling a blur-to-clear streaming effect.
(2) \textit{Global-diffusion} refinement formulates generation as multi-step denoising over the entire sequence, starting from noise or a coarse initialization and progressively refining to a complete output. This mechanism has been successfully adapted to both text~\cite{nie2025large, li2025lavida, song2025seed,li2022diffusionlm} and multimodal generation~\cite{xin2025lumina, yang2025mmada}.

\subsection{Efficient Streaming Generation}
\label{subsec:efficient_streaming_generation}

Given the extensive scope of LLM optimization, we narrow our focus strictly to the streaming process itself, analyzing decoding and memory efficiency.\footnote{We provide related survey papers on efficient LLMs in~\autoref{Appendix-A} for reference.} 
As \textit{token-wise} decoding remains dominant, we focus on its optimization for efficient streaming.

\paragraph{Decoding Path Acceleration}
To mitigate autoregressive latency, optimizations modify the execution trajectory along two dimensions.
(1) \textit{Token-path} methods generate parallel candidate chains to relax strict serial dependency, including multi-path and speculative decoding~\cite{leviathan2023specdec,xiao2024parallelspec}.
For instance, speculative decoding~\cite{chen2023accelerating,cai2024medusa,li2024eagle2} leverages a lightweight draft model to propose multiple candidate tokens in parallel, which are then verified and selectively accepted by a target model, reducing streaming latency.
(2) \textit{Layer-depth} methods adaptively shorten the network depth based on token difficulty~\cite{fan2024not,del2023skipdecode} . For instance, by employing layer skipping~\cite{zhao2025skipgpt}, models terminate the execution path prematurely.

\paragraph{Memory Efficiency}
Since the KV cache grows linearly, optimizations aim to decouple memory cost from generated length. \textit{Dynamic KV compression} methods limit the scope of attention targets during streaming decoding~\cite{liu2023scissorhands,zhang2023h2o,liao2025gkv,huang2024dynamic}. Representative implementations range from {sink-aware windowing}~\cite{xiao2023streamingllm}, which maintains a fixed budget for stability, to {dynamic decision strategies}~\cite{liao2025gkv} for KV cache management based on token importance.

\section{Sequential-Streaming LLMs: Processing Dynamic Input Streams}
\label{Sec:sequential-streaming}

Building upon the foundation of output-streaming, this section turns to sequential-streaming: the continuous perception of \textit{dynamic input streams}. The core technical imperative shifts from generation latency to sustainability. Specifically, we focus on two core mechanisms: handling incremental inputs to avoid re-computation, and optimizing context management to accommodate long input streams.

\subsection{Incremental Encoding}
\label{subsec:incremental-encoding}

Incremental encoding processes incoming streams solely based on past states, \textit{with historical representations remaining unchanged under subsequent streaming inputs}, avoiding quadratic re-computation. The central issue lies in \textit{how to define encoding units}, such that the encoding of each unit is not influenced by future information. Depending on the unit construction strategy, we categorize two types: \textit{atomic encoding} and \textit{fragmented encoding}.

\paragraph{Atomic Encoding}
This paradigm is applicable to streams that have inherent delimiters aligned with the model’s processing unit.
(1) \textit{Native Discrete Tokens}: Text is the primary example, where input is naturally segmented into discrete tokens whose representations remain unchanged as new tokens arrive~\cite{kudo2018subword,kudo2018sentencepiece}. 
(2) \textit{Pre-defined Units}: 
Certain modalities admit pre-defined atomic units independent of future context. For example, video streams can be incrementally processed at the frame level, where each frame serves as a fixed encoding unit and is encoded without being influenced by subsequent frames~\cite{dosovitskiy2020image,radford2021learning}.

\paragraph{Fragmented Encoding}
Fragmented encoding handles raw continuous signals (e.g., audio waveforms and video pixel streams) without natural delimiters by introducing artificial boundaries to interface with discrete LLM architectures. Boundary construction typically follows two strategies. (1) \textit{fixed-interval partitioning}, which slices streams at uniform temporal intervals for efficiency but may disrupt semantic units~\cite{wang2024simul,machavcek2023turning}. (2) \textit{semantic-driven partitioning}, which leverages content-dependent cues, such as  word boundaries in speech~\cite{zhang2023end,graves2012connectionist} and shot or scene transitions in video~\cite{zhao2023streaming}, to better preserve semantic coherence at higher computational cost.

\subsection{Streaming Context Management}
\label{subsec:context-management}
Streaming context management focuses on maintaining and updating contextual information during incremental processing under limited memory and computation budgets. It can be viewed through three complementary aspects: \textit{what information to keep} over long-running streams (memory), \textit{how to store and update} it across decoding steps (KV cache), and \textit{how to efficiently access} it via optimized attention mechanisms (attention).

\paragraph{Memory Retention}
Memory retention concerns what historical information should be preserved or discarded during long-running input streaming. We classify these methods into two primary categories.
(1) \textit{Salient content selection and eviction} approaches focuses on identifying and retaining salient tokens or segments while discarding less informative or redundant content as the stream grows~\cite{zhang2024flash,yao2025timechat,qian2024streaming,wang2025accelerating}. Selection criteria are typically based on importance estimation, recency, or task relevance, enabling bounded memory usage under continuous inputs.
(2) Instead of outright discarding past information, \textit{token merging and memory consolidation} compress historical representations by aggregating multiple tokens or states into more compact forms~\cite{zhong2024memorybank,wang2023augmenting,zeng2025streamforest,chen2025streamingtom}. Such strategies preserve coarse-grained contextual information while reducing memory footprint, allowing long-term context to be maintained in a compressed manner.

\paragraph{KV Cache and Attention Management}
While memory retention operates at the input level, this component focuses on the \textit{internal} maintenance of intermediate states and the optimization of attention computation. Since the attention range dictates which historical states are required for generation, attention access patterns and cache storage strategies are inherently coupled in streaming scenarios.
We categorize these strategies into two complementary directions.
(1) \textit{Attention-Aware Eviction}: These methods bound memory growth by restricting the attention mechanism to a sparse subset of historical tokens. By identifying and retaining only critical states, such as recent tokens maintained by a sliding window and high-importance attention sinks or heavy hitters, the model can safely evict unaccessed KV pairs, ensuring constant time and memory complexity without disrupting generation quality.
~\cite{xiao2023streamingllm,li2024snapkv,cai2024pyramidkv,yang2025streammem,liao2025gkv}.
(2) \textit{Representation Compression}: Complementary to eviction, compression approaches reduce the memory footprint of the \textit{retained} states. Techniques such as low-bit quantization or low-rank approximation compress the key--value representations, allowing the model to accommodate longer effective contexts within a fixed memory budget~\cite{liu2024kivi,hooper2024kvquant,he2024zipcache,liu2025chunkkv}.

\begin{figure}[t]
\begin{center}
\includegraphics[width=\linewidth]{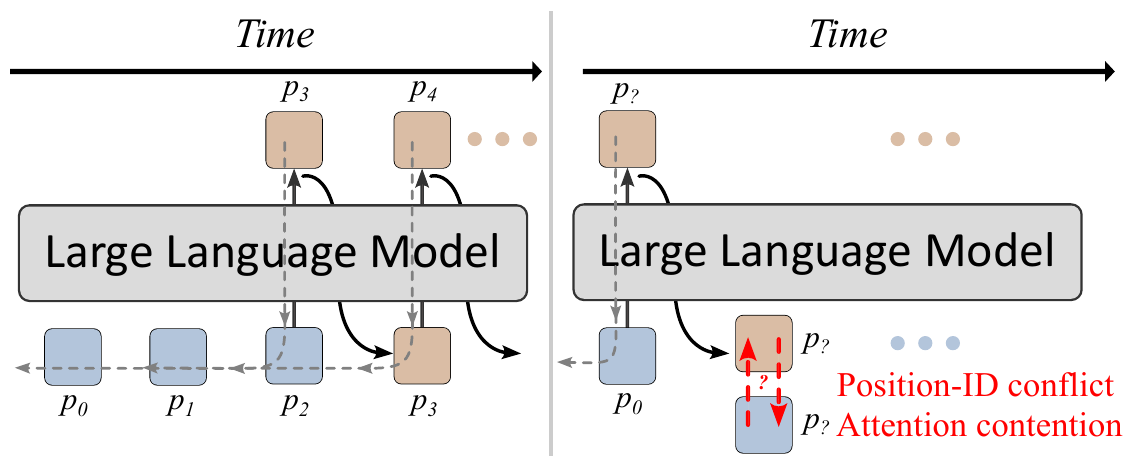}
\end{center}
\vspace{-1em}
\caption{\small Illustration of structural conflicts when adapting batch-oriented LLMs (left) to concurrent streaming (right), where \SolidArrow~indicates the token generation direction, \DashedArrow~denotes attention dependencies, \BlueBlock~blocks represent the input, and \OrangeBlock~blocks represent the output. (1) \textbf{Attention contention}: Ambiguous causal dependency between the newly inserted streaming input and historical outputs. (2) \textbf{Position-ID conflict}: The new streaming input and generated output compete for the identical position ID.}
\label{Fig:conflicts}
\end{figure}
\section{Concurrent-Streaming LLMs: The Streaming of Real-Time Interaction}
\label{Sec:Concurrent-streaming}

\begin{figure}[t]
\begin{center}
\includegraphics[width=0.9\linewidth]{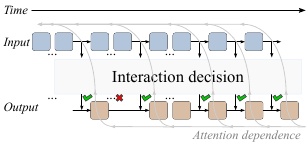}
\end{center}
\caption{\small Illustration of interaction decision in concurrent streaming LLMs, where the model learns to dynamically schedule reading inputs and emitting outputs.}
\label{Fig:interaction}
\end{figure}

Concurrent-streaming represent a crucial step toward real-time interactive intelligence, requiring LLMs to simultaneously process streaming inputs and generate outputs. However, this dynamic paradigm diverges from standard static pre-training. First, regarding architecture adaptation, concurrent streaming introduce structural conflicts, as illustrated in~\autoref{Fig:conflicts}. Second, synchronization control governs system interactivity by dynamically deciding when to alternate between reading and writing, balancing responsiveness and coherence, as illustrated in~\autoref{Fig:interaction}. Accordingly, we categorize existing research into \textit{architecture adaptation} and \textit{interaction policy}.

\subsection{Architecture Adaptation}
\label{subsec:architecutre_adaptation}

Architecture adaptation mitigates structural conflicts inherent in concurrent processing, including attention contention and positional conflicts (\autoref{Fig:conflicts}). Attention contention arises when continuously arriving inputs interleave with generation, making attention dependency ordering ambiguous, while positional conflicts occur when asynchronously injected inputs overlap with output positions. Existing work redesigns input–output interaction mechanisms, which we categorize into four representative streaming paradigms.

\paragraph{Re-encoded streaming}
The model re-encodes all historical caches whenever new input arrives~\cite{deng2025simuls2s,agostinelli2024simul,guo2024sillm}.
By recomputing representations over the entire context, this approach eliminates attention contention and positional misalignment, preserving batch-equivalent attention dependencies. However, the resulting computational overhead limits its applicability to long-context and real-time settings~\cite{guo2024decoder,raffel2024simultaneous}.

\begin{table*}[t]
\centering
\small

\renewcommand{\arraystretch}{1.15}
\setlength{\extrarowheight}{2pt}

\begin{tabular}{m{0.3cm}| m{3.4cm}| m{3.4cm}| m{3.4cm}| m{3.4cm}}
\hline
& \multicolumn{1}{c|}{\textbf{Re-encoded streaming}}
& \multicolumn{1}{c|}{\textbf{Concatenated streaming}}
& \multicolumn{1}{c|}{\textbf{Interleaved streaming}}
& \multicolumn{1}{c}{\textbf{Grouped streaming}} \\
\hline

\centering\rotatebox{90}{Illustration}
& \includegraphics[width=0.21\textwidth]{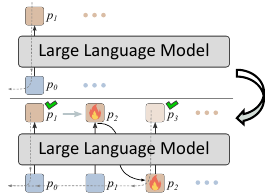}
& \includegraphics[width=0.21\textwidth]{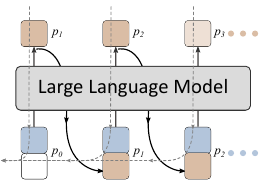}
& \includegraphics[width=0.21\textwidth]{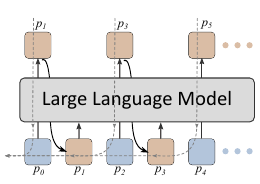}
& \includegraphics[width=0.21\textwidth]{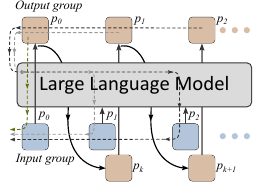}
\rule{0pt}{2.8ex} \\
\hline

\centering\rotatebox{90}{Attn.}
& Re-encode all past caches when new input arrives to match pretraining.
& Concatenate the input and output tokens into a composite token per step.
& Interleave input and output tokens on the timeline.
& Restrict attention within input and output groups to match pretraining. \\
\hline

\centering\rotatebox{90}{Pos.}
& Reassign positions via full re-encoding.
& Assign monotonic positions over concatenation.
& Assign positions by interleaved time order.
& Maintain separate positional spaces per group. \\
\hline
\end{tabular}

\caption{\small Comparison of concurrent-streaming architecture adaptation methods from the perspectives of attention (Attn.) and position (Pos.). \SolidArrow~indicates the token generation direction, while \DashedArrow~denotes attention dependencies. \BlueBlock~blocks represent the input stream, and \OrangeBlock~blocks represent the output stream. $p$ indicates the corresponding position ID.}
\label{tab:Paradigms}
\end{table*}

\paragraph{Concatenated streaming}
Concatenated streaming concatenates the newly arrived input tokens with the previously generated outputs and feeds them jointly into the model at each step~\cite{Qwen3-Omni,xu2025qwen2,ding2025kimi,shikhar2025llmvox}.
This design resolves both conflicts by unifying attention and positional ordering, but incurs growing memory and latency and requires architectural changes and retraining~\cite{shikhar2025llmvox}.

\paragraph{Interleaved streaming}
This paradigm interleaves input and output tokens within a shared sequence, assigning attention and positional encodings according to their temporal order~\cite{chen2024videollm,du2024cosyvoice,liu2025sync,xu2025streamingvlm,chen2025livecc,qian2025dispider}. It preserves the temporal flow of streaming interaction, enabling input and output to coexist with consistent ordering~\cite{chen2025livecc}. While balancing computational efficiency and real-time continuity, it requires synchronization mechanisms to prevent dependency leakage.

\paragraph{Grouped streaming}
Group streaming partitions input and output tokens into separate groups, each with independent attention relations and position IDs~\cite{liu2024streamchat,tong2025streamingthinker,tong2025llm,lin2026speak}. This design eliminates attention contention while maintaining isolated positional spaces, and empirical results show that grouped positional encoding preserves streaming performance and can improve parallelism and efficiency.

\subsection{Interaction policy}
\label{subsec:interaction_policy}
Interaction policy governs read–write synchronization in concurrent LLMs, balancing latency and output quality. Existing strategies fall into three paradigms based on their optimization approach: rule-based, SFT-based, and RL-based policies.

\paragraph{Rule-based Interaction}
Rule-based approaches rely on predetermined schedules or statistical thresholds, offering interpretability and control without requiring model parameter updates.
(1) \textit{Pre-defined} strategy enforce a rigid, content-agnostic read-write rhythm~\cite{ma2019stacl,chen2025livecc,tong2025streamingthinker}.
The most representative approach is the \textit{Wait-$k$ policy}~\cite{ma2019stacl}. In this strategy, the model always waits for $k$ tokens or segments of input lag before generating the corresponding output.
While efficient and easy to implement, pre-defined policies lack adaptability to varying input complexity and rate fluctuations.
(2) \textit{Adaptive thresholding} methods utilize real-time inference statistics as decision signals to improve flexibility~\cite{agostinelli2024simul,yang2025sasst}. 
These policies trigger read/write actions based on metric thresholds (e.g., attention weights) rather than a fixed schedule.
For instance, SimulS2S~\cite{agostinelli2024simul} monitors model confidence and pauses generation to read more context whenever uncertainty exceeds a safety margin, effectively adapting to the difficulty of the incoming stream.

\paragraph{SFT-based Interaction}
Moving beyond manual rules, supervised approaches leverage labeled data to explicitly train the model to predict the optimal interaction timing.
(1) \textit{In-context prediction} paradigm integrates decision-making directly into the autoregressive generation process~\cite{chen2024videollm,koshkin2024transllama}. Here, the LLM is fine-tuned to emit special control tokens (e.g., \texttt{<EOS>} or \texttt{<WAIT>} ) alongside standard text. This strategy unifies policy execution with language modeling, allowing the model to leverage its reasoning capabilities for control.
(2) \textit{Auxiliary decision} employ auxiliary decision modules to decouple control from generation~\cite{zhao2025drfrattn,chen2024divergence,qian2025dispider}. This typically involves training a lightweight classifier to output a binary decision. By isolating the interaction signal, this approach allows for focused supervision on the decision boundary without interfering with the semantic distribution of the generated text.

\begin{table}[t]
\centering
\small
\setlength{\tabcolsep}{3pt}
\resizebox{\linewidth}{!}{
\begin{tabular}{ccccc}
\toprule

\textit{Streaming-In} 
& \textit{Bound} 
& \textit{Inc.}
& \textit{Cxt.}
& \textit{Example methods}\\

\midrule

Text  & Memory & - & \blueyes & StreamingDialogue~\cite{machavcek2023turning}\\
Audio & Causal, Memory& \blueyes & \blueyes & WhisperStreaming~\cite{li2024streamingdialogue}\\
Video & Memory & - & \blueyes & Timechat-online~\cite{yao2025timechat}\\

\bottomrule
\end{tabular}
}
\caption{\small Summary of \textbf{sequential streaming} tasks. Incremental encoding (\textit{Inc}) and context management (\textit{Ctx}) are the key technical dimensions. The checkmark (\textcolor{myblue}{\ding{51}}) indicates the scope covered by existing research.}
\label{tab:tasks_sequential}
\vspace{-1em}
\end{table}

\begin{table*}[t]
\centering
\small
\setlength{\tabcolsep}{3pt}
\resizebox{\linewidth}{!}{
    \begin{tabular}{ccccccccccccccc}
        \toprule
        
        \multirow{2}{*}{\textit{Task type}} & \multirow{2}{*}{\textit{Level}}& \multicolumn{2}{c}{\textit{Modality}} & \multicolumn{4}{c}{\textit{Paradigm}} & \multicolumn{3}{c}{\textit{Interaction policy}} & \multirow{2}{*}{\textit{Example methods}}\\
        \cmidrule(lr){3-4} \cmidrule(lr){5-8} \cmidrule(lr){9-11}
        \multicolumn{2}{l}{} &\textit{In} &\textit{Out} & \textit{R.} & \textit{C.} & \textit{I.} & \textit{G.} &\textit{Rule} &\textit{SFT} &\textit{RL} & \\ 
        
        \midrule
        Translation & $\mathcal{X} \to \mathcal{Y}$       & T/S & T/S     & \blueyes & \blueyes & \blueyes & \blueyes & \blueyes & \blueyes & \blueyes &  Seed LiveInterpret 2.0~\cite{cheng2025seed}\\
        
        Detection & $\mathcal{X} \to \mathcal{Y}$         & T/S/V  &T     & - & - & \blueyes & - & - & \blueyes & - & FineHarm~\cite{li2025judgment}\\
        
        ASR & $\mathcal{X} \to \mathcal{Y}$               & S  &T         & - & - & \blueyes & \blueyes & \blueyes & \blueyes & - & ReaLLM~\cite{seide2024speech}, Llama-omni~\cite{fang2024llama}\\
        
        TTS & $\mathcal{X} \to \mathcal{Y}$               & T  &S         & - & \blueyes & - & \blueyes & \blueyes & \blueyes & - & Cosyvoice~\cite{du2024cosyvoice}, DSM~\cite{zeghidour2025streaming}\\
        
        QA & $\mathcal{X} \to \mathcal{Y}$                & T/S/V & T/S   & - & \blueyes & \blueyes & - & - & \blueyes & \blueyes & Qwen3-omni~\cite{Qwen3-Omni}, VideoLLM-online~\cite{chen2024videollm}\\
        
        Description & $\mathcal{X} \to \mathcal{Y}$       & V  &T         & - & - & \blueyes & \blueyes & \blueyes & \blueyes & - & LiveCC~\cite{chen2025livecc}, StreamMind~\cite{ding2025streammind}\\
        
        VLA & $\mathcal{X} \to \mathcal{Y}$               & V  &T         & - & - & \blueyes & - & \blueyes & - & - & StreamVLN~\cite{wei2025streamvln}, ActiveVLN~\cite{zhang2025activevln}\\
        
        \hdashline
        Reasoning & $\mathcal{X} \to \mathcal{Z}$         & T/S/V  &T/S   & - & - & \blueyes & \blueyes & \blueyes & - & - & StreamingThinker~\cite{tong2025streamingthinker}\\
        
                  & $\mathcal{Z} \to \mathcal{Y}$         & T/S/V  &T/S   & - & - & \blueyes & - & \blueyes & - & - & AsyncReasoning~\cite{yakushev2025asynchronous}\\
                  
        \hdashline     
        Tool usage & $\mathcal{X} \to \mathcal{Z}$        & T/S/V  &T     & - & - & \blueyes & - & \blueyes & - & - &AViLA~\cite{zhang2025avila}, StreamRAG~\cite{arora2025stream}\\
                  & $\mathcal{Z} \to \mathcal{Y}$         & T/S/V  &T   & - & - & \blueyes & - & \blueyes & - & - & Conveyor~\cite{xu2024conveyor}, AsyncLM~\cite{gim2024asynchronous}\\
        
        \bottomrule
    \end{tabular}
}
\caption{\small Summary of \textbf{concurrent streaming} tasks and representative methods. Tasks are categorized by processing depth (\textit{Level}), where $\mathcal{X} \to \mathcal{Y}$ denotes direct mapping (perception) and $\mathcal{X} \to \mathcal{Z} \to \mathcal{Y}$ denotes intermediate processing with a latent state $\mathcal{Z}$ (cognition). Modality: text (\textit{T}), speech (\textit{S}), vision (\textit{V}). Streaming Paradigm: re-encoding (\textit{R}), Concatenated (\textit{C}), Interactive (\textit{I}), Group (\textit{G}). Interaction Policy: Rule-based (\textit{Rule}), SFT-based (\textit{SFT}), and RL-based (\textit{RL}). The checkmark (\textcolor{myblue}{\ding{51}}) indicates the scope covered by existing research.}
\label{tab:tasks_concurrent}
\end{table*}

\paragraph{RL-based Interaction}
RL-based policies model interaction control as sequential decision-making, where the LLMs selects read or write actions based on the current context~\cite{wang2025mmduet2,cheng2025seed,xu2025seqpo}. Optimizing quality–latency rewards enables the discovery of non-trivial interaction patterns that are difficult to encode with static rules.
For example, MMDuet2~\cite{wang2025mmduet2} formulates proactive video interaction as an RL-driven control problem, enabling asynchronous perception and reaction under streaming video inputs.

\section{Streaming Applications and Tasks}
This section reviews the application-level tasks enabled by streaming LLMs, building upon the methodological taxonomy established in Sections 3–5. 
Notably, since output streaming is a universal property of LLM-based generation, we concentrate on task settings where streaming arises from incremental input, real-time interaction, or bidirectional coupling between input and output.

\paragraph{Sequential Streaming Tasks}
Sequential streaming tasks target long, unbounded input streams that cannot be processed in a single pass due to resource limitations. For instance, streaming long video understanding~\cite{zhang2024flash, yao2025timechat} requires incremental video encoding, followed by immediate decoding upon query arrival.
As summarized in~\autoref{tab:tasks_sequential}, different modalities emphasize distinct technical components.

\paragraph{Concurrent Streaming Tasks}

Concurrent streaming covers multimodal tasks that require simultaneous input reception and output generation. Based on processing depth, these tasks can be divided into two levels.
(1) Perception-Level ($\mathcal{X} \rightarrow \mathcal{Y}$): Models focus on direct cross-modal mappings with minimal latency, including streaming translation (e.g., Seed LiveInterpret 2.0~\cite{cheng2025seed}), ASR/TTS (e.g., CosyVoice~\cite{du2024cosyvoice}), real-time video captioning (e.g., LiveCC~\cite{chen2025livecc}), and streaming QA (e.g., Qwen3-Omni~\cite{Qwen3-Omni}).
(2) Cognition-Level ($\mathcal{X} \rightarrow \mathcal{Z} \rightarrow \mathcal{Y}$): Tasks require maintaining and updating a latent state $\mathcal{Z}$ to support complex behaviors such as streaming reasoning (e.g., StreamingThinker~\cite{tong2025streamingthinker}) and streaming tool usage (e.g., AViLA~\cite{zhang2025avila}). Here, the latent state decouples immediate perception from final output generation.
We summarize the corresponding technical categories of these tasks in~\autoref{tab:tasks_concurrent}.

\section{Future Directions}
\vspace{-0.5em}
To provide a comprehensive roadmap, we categorize future research into two complementary perspectives: the \textit{technical level} (i.e., how to build better streaming models) and the \textit{application level} (i.e., how to apply streaming models).

\paragraph{Technical Level}
(1) \textit{\textbf{Efficient Streaming LLMs}}. Efficiency under strict latency and memory constraints remains a core challenge, involving incremental encoding, decoding acceleration, and long-term context management. 
(2) \textit{\textbf{Alternative Concurrent Streaming Paradigms}}. Beyond interleaved and group-based strategies, more effective streaming paradigms remain to be explored. In particular, extending streaming interaction to semi-autoregressive or block-wise generation frameworks presents a promising yet underexplored direction. 
(3) \textit{\textbf{Proactive Interaction Policies}}. Designing interaction policies that adaptively balance reading and generation is essential for real-time streaming performance.
(4) \textit{\textbf{Interpretability}}. The behavioral dynamics of LLMs in interactive streaming settings remain largely unexplored, calling for greater interpretability.

\paragraph{Application Level}
(1) \textit{\textbf{Expansion of Streaming Modalities}}. Current streaming LLMs primarily focus on text, audio, and basic video interactions. Extending streaming LLMs to additional modalities requires transcending these limitations toward complex, omni-modal continuous streams (e.g., parallel video-audio streams) to achieve real-time streaming multimodal understanding and generation in highly dynamic environments. 
(2) \textit{\textbf{Expansion of Concurrency Levels}}. A promising direction is to expand current streaming LLMs from two-level perceptual concurrency (e.g., ``listen-while-speaking'' and ``read-while-thinking'') to deeper, multi-level asynchronous processing. This includes 3-level streaming (introducing streaming ``perceiving, reasoning, and generation'') and 4-level streaming (introducing concurrent ``perceiving, reasoning, tool-using, and generation'') to achieve true \textit{multi-stream intelligence}. 
(3) \textit{\textbf{Expansion of Streaming Tasks}}. The application of streaming LLMs is expected to shift from simple, passive responses toward complex proactive interactions and long-context engagements. Advancing these capabilities involves empowering models to actively initiate interventions and maintain long-term memory, ultimately achieving brain-like streaming intelligence.

\section{Conclusion}

This survey presents a unified view of streaming LLMs by clarifying their definitions and organizing existing approaches into output-streaming, sequential-streaming, and concurrent-streaming paradigms based on data flow and interaction concurrency. We review representative methodologies and application scenarios, and discuss the fundamental challenges posed by real-time and interactive settings. We hope this work serves as a concise reference and a conceptual foundation for future research on streaming intelligence.

\section*{Limitations}
This survey focuses on clarifying the conceptual landscape of Streaming Large Language Models through unified definitions, paradigms, and representative methods. As a result, it does not aim to provide an exhaustive comparison of all existing implementations or a comprehensive empirical evaluation across tasks and systems.
Moreover, our discussion primarily centers on high-level design principles and paradigms, leaving detailed system-level optimizations and deployment-specific considerations for future studies.

\bibliography{reference}

\clearpage
\setcounter{figure}{0}   
\setcounter{table}{0}
\setcounter{equation}{0}
\appendix

\section{Survey Scope and Positioning}
\label{Appendix-A}

\subsection{Motivation and Necessity of This Survey}
The motivation for this survey stems from three key observations regarding the current landscape of Large Language Models (LLMs): the paradigm shift to streaming scenarios, ambiguity in "streaming" terminology, and absence of comprehensive reviews in streaming LLMs domain.

\paragraph{The Paradigm Shift to Streaming Scenarios}
While LLMs have demonstrated remarkable capabilities across various static inputs, real-world deployment increasingly demands streaming interaction. Applications such as digital human assistants, real-time simultaneous interpretation, and embodied robotics require models to process continuous input streams and generate low-latency responses. The transition from "static batch processing" to "dynamic streaming interaction" presents unique challenges in memory management, temporal coherency, and inference efficiency that traditional LLM research overlooks.

\paragraph{Ambiguity in "Streaming" Terminology}
There is currently a significant semantic ambiguity in the usage of the term "streaming" within the community. It is often conflated across three distinct dimensions: streaming generation (token-by-token output), streaming processing (handling dynamic input context), and streaming interaction (dynamic generate with partial and dynamic input). This survey aims to disambiguate these concepts and provide a rigorous taxonomy.

\paragraph{Absence of Comprehensive Reviews}
Despite the surge in related research, there is a notable lack of a systematic survey dedicated to Streaming LLMs.

\subsection{Focus and Scope Delimitation}
To ensure depth and coherence, we delineate the scope of this survey as follows: 
We primarily focus on decoder-only LLMs, and we structure the survey by tracing the evolution of streaming capabilities: from static input / streaming output (standard generation), to streaming Input / streaming output (infinite context processing), and finally to dynamic interaction (duplex/omni-streaming). 
We conducted a systematic literature review of top-tier venues in AI, NLP, CV, and speech, \textbf{with a cutoff date of December 2025}.

\subsection{Comparison with Existing Surveys}
\label{subsec:comparison_with_existing_surveys}

While our survey establishes a unified taxonomy for \textit{Streaming LLMs} centered on dynamic data flow and real-time interaction, it is crucial to delineate its scope from other prominent research directions in the LLM landscape. Below, we contrast our focus with three major categories of existing surveys: Efficient LLMs, Multimodal LLMs, and Long-Context LLMs.

\paragraph{Efficient LLMs.}
The technologies surveyed under the field of Efficient LLMs, including model compression and KV cache management, are foundational technology for efficient, accurate and intelligent Streaming LLMs \cite{li2024survey, dantas2025review, cheng2025survey}. However, existing typical surveys in this category, such as the survey on KV cache management for acceleration \cite{li2024survey}, the comprehensive survey on efficient LLMs \cite{dantas2025review}, and the review on compression techniques \cite{cheng2025survey}, predominantly analyze these methods from an offline and static perspective. Central questions of these surveys is mostly on how to reduce the computational or memory footprint of a model that is operating on a complete, existing context to acquire higher throughput or enable development on hardware with constrained resources. In comparison, this survey re-contextualizes these optimizations  within a streaming paradigm. Techniques such as dynamic KV cache management and lightweight model adaptation under the overarching imperative of online, real-time interaction are unified. The key challenge shifts from static resource reduction to dynamic runtime budgeting under the strict latency constraints of streaming, where inputs are incrementally available, as is concurrent streaming defined, and outputs must also be generated incremenntally. Thus, while efficient LLM research only casts light on how can we run the model more efficiently, this research also  asks how can it read, listen, see and respond efficiently as the world unfolds.

\paragraph{Multimodal LLMs.}
The field of MLLMs~\cite{zhang2024mm} focuses on augmenting language models with the ability to process and generate content across diverse modalities like vision, audio, and video. Key challenges include cross-modal alignment, fusion strategies, and the design of modality-specific encoders and decoders. Although some MM-LLM applications (e.g., real-time video analysis or speech-to-speech translation) are inherently streaming, the primary goal of MM-LLM research is to achieve strong performance on multimodal understanding and generation benchmarks. Our survey, however, abstracts away from the specifics of any single modality. We treat the input and output as generic token streams and instead concentrate on the \textit{temporal dynamics of the interaction}. A streaming LLM architecture, as defined in our work, can serve as the backbone for a multimodal system, but the core innovations we survey—such as concurrent perception-generation loops and infinite context processing—are orthogonal to the problem of modality grounding. Our focus is on \textit{how} information flows over time, not \textit{what} the information represents.

\paragraph{Long-Context LLMs.}
Surveys in this category, such as \cite{liu2025comprehensive} and \cite{wang2024beyond}, primarily focus on expanding the model's static capacity to process extremely long, finite input sequences (e.g., long documents or multi-turn histories). Their core goal is to extend the usable context window and make inference over long sequences efficient, covering key technologies like positional encoding extrapolation, efficient attention architectures (e.g., sparse attention), and sophisticated KV-cache management. While these advances in long-context modeling provide a crucial foundational capability for processing extensive information, their perspective is largely centered on a "read-then-write" inference paradigm for offline, bounded inputs. In stark contrast, our survey on Streaming LLMs investigates the dynamic interaction paradigm required for unbounded, real-time token streams. We focus on the unique challenges of concurrent reading and writing, incremental processing of growing states, and online context/KV budgeting under strict latency constraints. Therefore, while long-context techniques are often essential enabling components, our work shifts the focus from merely enlarging a fixed context window to orchestrating continuous, low-latency reasoning and generation within an ever-flowing data stream.

\begin{table*}[t]
    \centering
    \small
    \renewcommand{\arraystretch}{1.6}
    \newlength{\firstcolwidth}
    \settowidth{\firstcolwidth}{\cite{caffagni2024revolution}}
    \begin{tabularx}{\textwidth}{@{} 
        >{\raggedright}p{\firstcolwidth} 
        >{\raggedright}p{2.8cm}
        >{\raggedright\arraybackslash}X
        >{\raggedright\arraybackslash}X 
        @{}}
        \toprule
        \textbf{Typical Surveys} & 
        \textbf{Primary Focus} & 
        \textbf{Typical Technologies Covered} & 
        \textbf{Differentiation in This Survey} \\
        \midrule
        \rowcolor{gray!15} \multicolumn{4}{c}{\textit{Survey Category: Efficient LLMs}} \\
        \cite{li2024survey} \\
        \cite{dantas2025review} \\
        \cite{cheng2025survey} \\
        \cite{Wu2026survey}
        & Compression/adaptation and memory bottlenecks.         
        & 1) Compression: quantization, pruning, distillation, low-rank; and 2) KV-cache management: selection / eviction, cache compression, offloading, sliding-window / hierarchical cache.
        & Prior surveys treat compression and KV-cache optimization as separate threads; we unify them under streaming interaction, highlighting online constraints and dynamic runtime budgeting. \\
        \cmidrule(r){1-4}
        \rowcolor{gray!15} \multicolumn{4}{c}{\textit{Survey Category: Multimodal LLMs}} \\
        \cite{zhang2024mm} \\
        \cite{caffagni2024revolution}
        & Architectures, training recipes, and benchmarks for MLLMs.
        & 1) Encoder + Projector + LLM, alignment module, tokenizer; and 2) multimodal pretraining \& instruction tuning.
        & Prior MLLM surveys assume fixed inputs and emphasize alignment and benchmarked capabilities. We focus on streaming interaction with token stream abstraction, concurrent IO, incremental perception, and online memory and budget control. \\
        \cmidrule(r){1-4}
        \rowcolor{gray!15} \multicolumn{4}{c}{\textit{Survey Category: Long-Context LLMs}} \\
        \cite{wang2024beyond} \\
        \cite{liu2025comprehensive}
        & Long-context modeling: extending usable context windows and making long-sequence inference efficient.
        & 1) Position extrapolation / interpolation; 2) efficient long-sequence attention and architectures; 3) KV-cache management (compression, eviction, and offloading); and 4) workflow-level augmentation (prompt compression, retrieval/external memory).
        & Prior surveys focus on enlarging a fixed context window for offline inputs or read then write inference. We study streaming token streams with concurrent read and write, incremental inputs, growing states, and online context and KV budgeting for unbounded streams. \\
        \bottomrule
    \end{tabularx}
    \caption{Comparison between this survey and existing related surveys. We highlight the unique positioning of our work in the context of streaming interaction.}
    \label{tab:survey_comparison}
\end{table*}
\section{Supplementary Literature}
\label{Appendix-B:literature}
Due to space limitations, we defer a broader collection of related work to this appendix. Following the taxonomy in Figure~\ref{fig:taxo_of_streaming_llms}, we organize additional literature into three paradigms: (1) Output-streaming LLMs, (2) Sequential-streaming LLMs, and (3) Concurrent-streaming LLMs. This appendix complements the main text by summarizing representative yet less-discussed threads and implementations, rather than aiming for an exhaustive bibliography.

Table~\ref{tab:appendix-B-out} presents additional methods for output-streaming LLMs, organized by streaming generation mechanisms and efficiency techniques.

\begin{table*}[t]
\centering
\small
\setlength{\tabcolsep}{2.5pt}
\renewcommand{\arraystretch}{1.25}
\resizebox{\linewidth}{!}{
\begin{tabular}{
    >{\centering\arraybackslash}p{1.1cm}
    >{\centering\arraybackslash}p{1.1cm}
    >{\centering\arraybackslash}p{1.4cm}
    >{\centering\arraybackslash}p{1.6cm}
    >{\raggedright\arraybackslash}p{10cm}
}
\toprule
\rowcolor{gray!15} \multicolumn{5}{c}{\textbf{\textit{Streaming Generation}}} \\
\midrule
\multicolumn{3}{c}{\textit{Mechanism}} & \multirow{2}{*}{\textit{Modality-Out}} & \multicolumn{1}{c}{\multirow{2}{*}{\textit{Methods}}} \\
\cmidrule(lr){1-3}
\textit{Token} & \textit{Block} & \textit{Refinement} & & \\
\midrule
\blueyes & - & - & T & GPT~\cite{OpenAI2023}, Gemini~\cite{team2023gemini}, Qwen3~\cite{yang2025qwen3}, DeepSeek-V3~\cite{deepseekv3}, InternVL~\cite{chen2024internvl}, ChatGLM~\cite{glm2024chatglm}, Gemma~\cite{gemma} \\[0.25em]

\blueyes & - & - & S & AudioLM~\cite{borsos2023audiolm}, SpeechGPT~\cite{zhang2023speechgpt}, AudioPaLM~\cite{rubenstein2023audiopalm}, FireRedTTS~\cite{guo2024fireredtts}, Moshi~\cite{defossez2024moshi}, Llama-omni2~\cite{fang2025llama}, Qwen3-Omni~\cite{Qwen3-Omni}, StyLLE~\cite{haostylle}, Llmvox~\cite{shikhar2025llmvox}, SpeakStream~\cite{bai2025speakstream} \\[0.25em]

\blueyes & - & - & V & DALLE~\cite{ramesh2021dalle}, VideoPoet~\cite{kondratyuk2024videopoet}, Chameleon~\cite{chameleonteam2024chameleon}, Emu3~\cite{wang2024emu3}, Anole~\cite{chern2024anole}, Lumina-mGPT2.0~\cite{xin2025lumina}, Infinity~\cite{han2025infinity} \\[0.25em]

- & \blueyes & - & T & SAT~\cite{wang2018sat}, SoT~\cite{ning2023skeleton}, CtrlDiff~\cite{huang2025ctrldiff}, PredSent~\cite{hwang2025predsent}, Falcon~\cite{gao2025falcon}, SSD-LM~\cite{han2023ssd}, WeDLM~\cite{liu2025WeDLM}, Next-Block~\cite{tian2025next}, Block Diffusion~\cite{arriola2025block} \\[0.25em]

- & \blueyes & - & S & PALLE~\cite{yang2025pseudo}, SyncSpeech~\cite{sheng2025syncspeech}, DCAR~\cite{li2025robust}, StreamFlow~\cite{guo2025streamflow}, TtT~\cite{liu2025text}, DiTAR~\cite{jia2025ditar} \\[0.25em]

- & \blueyes & - & V & show-o~\cite{xie2024show}, XTRA~\cite{amrani2025sample}, NTP~\cite{ren2025next}, CausVid~\cite{yin2025slow}, BlockVid~\cite{zhang2025blockvid}, NBP~\cite{ren2025next}\\[0.25em]

- & - & \blueyes & T & Mask-Predict~\cite{ghazvininejad2019maskpredict}, LevT~\cite{gu2019levenshtein}, Insertion-Deletion~\cite{ruis2020insertiondeletion}, Diffusion-LM~\cite{li2022diffusionlm}, DiffuSeq~\cite{gong2022diffuseq}, D3PM~\cite{austin2021d3pm} \\[0.25em]

- & - & \blueyes & S & SoundStorm~\cite{borsos2023soundstorm}, Voicebox~\cite{le2023voicebox}, Specmaskgit~\cite{comunita2024specmaskgit}, IMPACT~\cite{huang2025impact}, Maskgct~\cite{wang2024maskgct}, DDM-TASTE~\cite{ku2025discretediffusionspeechtokens} \\[0.25em]

- & - & \blueyes & V & MaskGIT~\cite{chang2022maskgit}, Muse~\cite{chang2023muse}, DiT~\cite{peebles2022dit}, VAR~\cite{tian2024visual}, DetailFlow~\cite{liu2025detailflow}, DC-AR~\cite{wu2025dc} \\[0.4em]

\midrule
\rowcolor{gray!15} \multicolumn{5}{c}{\textbf{\textit{Streaming Efficiency}}} \\
\midrule
\multicolumn{2}{c}{\textit{Efficient}} & \multicolumn{1}{c}{} & \multirow{2}{*}{\textit{Modality-Out}} & \multicolumn{1}{c}{\multirow{2}{*}{\textit{Methods}}}  \\
\cmidrule(lr){1-2}
\textit{Decode} & \textit{Memory} & & & \\
\midrule
\blueyes & - & & T & Speculative Sampling~\cite{chen2023accelerating}, Medusa~\cite{cai2024medusa}, EAGLE2~\cite{li2024eagle2}, BiLd~\cite{kim2023speculative}, CTC-based Drafting~\cite{wen2024speculative}, FLY~\cite{liu2025drop}, SkipDecode~\cite{del2023skipdecode}, SkipGPT~\cite{zhao2025skipgpt}, EESD~\cite{liu2024speculative}, HiDrop~\cite{Wu2026HiDrop}, Visipruner~\cite{fan2025visipruner} \\[0.25em]

\blueyes & - & & S & LiveSpeech~\cite{dang2024livespeech}, MTP-SpecDec~\cite{nguyen2025accelerating}, SSD~\cite{lin2025accelerating}, VocalNet~\cite{wang2025vocalnet}, VADUSA~\cite{li2025fast}, PCG~\cite{yanuka2025principled} \\[0.25em]

\blueyes & - & & V & SJD~\cite{teng2024accelerating}, CSpD~\cite{wang2024continuous}, GSD~\cite{so2025grouped}, VVS~\cite{dong2025vvs}, FreqExit~\cite{li2025freqexit}, SkipVAR~\cite{li2025skipvar}, PAR~\cite{wang2025parallelized}, ADT-Tree~\cite{lei2025fast}, Lantern~\cite{jang2024lantern} \\[0.25em]

- & \blueyes & & T & StreamingLLM~\cite{xiao2023streamingllm}, H2O~\cite{zhang2023h2o}, Scissorhands~\cite{liu2023scissorhands}, Snapkv~\cite{li2024snapkv}, Dynamickv~\cite{zhou2024dynamickv}, Chunkkv~\cite{liu2025chunkkv} \\[0.25em]

- & \blueyes & & S & wu2024ts3~\cite{wu2024ts3}, LST~\cite{lu2025latent}, SpeechTokenPrediction~\cite{liu2025speech} \\[0.25em]

- & \blueyes & & V & HACK~\cite{qin2025head}, ScaleKV~\cite{li2025memory}, AMS-KV~\cite{xu2025ams}, LineAR~\cite{qin2025autoregressive} \\
\bottomrule
\end{tabular}
  }
\caption{Summary of additional literature on output-streaming LLMs, complementing the discussion in Sec.~\ref{Sec:output-streaming}.}
\label{tab:appendix-B-out}
\end{table*}

Table~\ref{tab:appendix-B-seq} summarizes methods for sequential-streaming LLMs, focusing on incremental encoding and streaming context management.

\begin{table*}[t]
\centering
\small
\setlength{\tabcolsep}{3pt}
\renewcommand{\arraystretch}{1.3}
 \resizebox{\linewidth}{!}{
\begin{tabular}{
    >{\centering\arraybackslash}p{2cm}
    >{\centering\arraybackslash}p{2cm}
    >{\centering\arraybackslash}p{1.8cm}
    >{\raggedright\arraybackslash}p{10.5cm}
}
\toprule
\rowcolor{gray!15} \multicolumn{4}{c}{\textbf{\textit{Incremental Encoding}}} \\
\midrule
\multicolumn{2}{c}{\textit{Type}} & \multirow{2}{*}{\textit{Modality-In}} & \multicolumn{1}{c}{\multirow{2}{*}{\textit{Methods}}} \\
\cmidrule(lr){1-2}
\textit{Fragmented Encoding} & \textit{Atomic Encoding} & & \\ 
\midrule
\blueyes & - & T & SimulMT~\cite{wang2024simultaneous}, Moshi~\cite{defossez2024moshi}, Codec~\cite{ye2025codec}, dmel~\cite{bai2024dmel}, Lightweight Audio Segmentation~\cite{frohmann2024segment}, Semantic VAD~\cite{shi2023semantic} \\[0.3em]

\blueyes & - & S &  Whisper-Streaming~\cite{machavcek2023turning}, SimulST~\cite{zhang2023end}, CTC~\cite{graves2012connectionist}, Speechtokenizer~\cite{xin2024speechtokenizer}, Moshi~\cite{defossez2024moshi}, Codec~\cite{ye2025codec}, dmel~\cite{bai2024dmel}, Lightweight Audio Segmentation~\cite{frohmann2024segment}, Semantic VAD~\cite{shi2023semantic} \\[0.3em]

\blueyes & - & V &  S-ViT~\cite{zhao2023streaming} \\[0.3em]

- & \blueyes & T & SaT~\cite{frohmann2024segment}, SegFree~\cite{iranzo2024segmentation}, WtP~\cite{minixhofer2023s}, subword regularization~\cite{kudo2018subword}, SentencePiece ~\cite{kudo2018sentencepiece},\\[0.3em]

- & \blueyes & V & ViT~\cite{dosovitskiy2020image}, CLIP~\cite{radford2021learning}\\

\midrule
\rowcolor{gray!15} \multicolumn{4}{c}{\textbf{\textit{Streaming Context Management}}} \\
\midrule
\multicolumn{3}{c}{\textit{Type}} & \multicolumn{1}{c}{\multirow{2}{*}{\textit{Methods}}} \\
\cmidrule(lr){1-3}
\textit{Mem.} & \textit{KV} & \textit{Attn.} & \\ 
\midrule
\blueyes & - & - & StreamingTOM~\cite{chen2025streamingtom}, MemoryBank~\cite{zhong2024memorybank}, LongMem~\cite{wang2023augmenting}, VideoStreaming~\cite{qian2024streaming}, Timechat-online~\cite{yao2025timechat}, Prunevid~\cite{huang2025prunevid}, DyCoke~\cite{tao2025dycoke}, ProVideLLM~\cite{chatterjee2025streaming}, VideoLLaMB~\cite{wang2025videollamb}, STREAMMIND~\cite{ding2025streammind}, VideoStreaming~\cite{qian2024streaming}, StreamingAssistant~\cite{jin2025streamingassistant}, Focus~\cite{wei2025focus}, StreamForest~\cite{zeng2025streamforest}, Flash-vstream~\cite{zhang2024flash} \\[0.3em]

- & \blueyes & - & H2o~\cite{zhang2023h2o}, PyramidKV~\cite{cai2024pyramidkv}, SnapKV~\cite{li2024snapkv}, StreamKV~\cite{chen2025streamkv}, STC~\cite{wang2025accelerating}, Streammem~\cite{yang2025streammem}, AViLA~\cite{zhang2025avila}, StreamingVLM~\cite{xu2025streamingvlm}, PyramidInfer~\cite{yang2024pyramidinfer}, DynamicKV~\cite{zhou2024dynamickv}, PrefixKV~\cite{wang2024prefixkv}, CAKE~\cite{qin2025cake}, SimLayerKV~\cite{zhang2024simlayerkv}, AdaKV~\cite{feng2024ada}, CriticalKV~\cite{feng2025identify}, LeanKV~\cite{zhang2024unifying}, RazorAttention~\cite{tang2024razorattention}, HeadKV~\cite{fu2024not}, DuoAttention~\cite{xiao2024duoattention} \\[0.3em]

- & - & \blueyes & Attention Sink~\cite{xiao2023streamingllm}, Sirllm~\cite{yao2024sirllm}, GLA~\cite{yang2023gated}, DeltaNet~\cite{yang2024parallelizing}, Lightning attention-2~\cite{qin2024lightning}, SAMPLEATTENTION~\cite{zhu2025sampleattention}, Lserve~\cite{yang2025lserve}, DCA~\cite{an2024training} \\

\bottomrule
\end{tabular}
}
\caption{Summary of additional literature on sequential-streaming LLMs, complementing the discussion in Sec.~\ref{Sec:sequential-streaming}.}
\label{tab:appendix-B-seq}
\end{table*}

\begin{table*}[t]
\centering
\small
\setlength{\tabcolsep}{3pt}
\resizebox{\linewidth}{!}{
    \begin{tabular}{
    >{\centering\arraybackslash}p{0.8cm}
    >{\centering\arraybackslash}p{0.8cm}
    >{\centering\arraybackslash}p{0.8cm}
    >{\centering\arraybackslash}p{0.8cm}
    >{\centering\arraybackslash}p{1cm}
    >{\centering\arraybackslash}p{1cm}
    >{\raggedright\arraybackslash}p{10cm}
    }
        \toprule
         \rowcolor{gray!15} \multicolumn{7}{c}{\textit{Streaming Paradigm}} \\
         \midrule
        \multicolumn{4}{c}{\textit{Paradigm}} & \multicolumn{2}{c}{\textit{Modality}} & \multicolumn{1}{c}{\multirow{2}{*}{\textit{Methods}}} \\
        \cmidrule(lr){1-4} \cmidrule(lr){5-6}
        \textit{R.} & \textit{C.} & \textit{I.} & \textit{G.} & \textit{In} &\textit{Out} & \\ 
        \midrule

        \blueyes & -  & -  &  - & T  &  T  & Simul-LLM~\cite{agostinelli2024simul}, SiLLM~\cite{guo2024sillm}, TransLLaMA~\cite{koshkin2024transllama}, CAST~\cite{koshkin2024llms}, RALCP~\cite{wang2024simultaneous}\\
        \blueyes & -  & -  &  - & S  &  T  & CAST~\cite{koshkin2024llms}, TransLLaMA~\cite{koshkin2024transllama}\\

        - & \blueyes  & -  &  - & T  &  S  & LLMVoX~\cite{shikhar2025llmvox}, Mini-Omni~\cite{xie2024mini}\\
        - & \blueyes  & -  &  - & S  &  S  & Mini-Omni~\cite{xie2024mini}\\
        - & \blueyes  & -  &  - & V  &  T  & ViSpeak~\cite{fu2025vispeak}\\

        - & -  & \blueyes  &  - & T  &  T  & EAST~\cite{fu2025llms}, Shanks~\cite{chiang2025shanks}\\
        - & -  & \blueyes  &  - & T  &  S  & STITCH~\cite{chiang2025stitch}\\
        - & -  & \blueyes  &  - & S  &  T  & EASiST~\cite{fu2025efficient}, InfiniSST~\cite{ouyang2025infinisstsimultaneoustranslationunbounded}, SASST~\cite{yang2025sasst}, StreamingASR~\cite{wan2026streaming}\\
        - & -  & \blueyes  &  - & S  &  S  & SALMONN-omni~\cite{zhang2024salmonnomni}\\
        - & -  & \blueyes  &  - & V  &  T  & Videollm-online~\cite{chen2024videollm}, LiveCC~\cite{chen2025livecc},  ProVideLLM~\cite{chatterjee2025streaming}, StreamBridge~\cite{wang2025streambridge}, LiveStar~\cite{yang2025livestar}, SVBench~\cite{yang2025svbench}, ProASIST~\cite{zhang2025proactive}\\

        - & -  & -  &  \blueyes & T  &  T  & StreamingGPE~\cite{tong2025llm}, StreamingThinker~\cite{tong2025streamingthinker}, DST~\cite{guo2024decoder}\\
        - & -  & -  &  \blueyes & S  &  T  &  StreamingGPE~\cite{tong2025llm}\\
        - & -  & -  &  \blueyes & V  &  T  & StreamChat~\cite{liu2024streamchat}, Speak-While-Watching~\cite{lin2026speak}, TaYS~\cite{Zhang2026TaYS}\\

        \midrule
        \rowcolor{gray!15} \multicolumn{7}{c}{\textit{Interaction Policy}} \\
        \midrule
        \multicolumn{3}{c}{\textit{Policy}} & \multicolumn{1}{c}{} & \multicolumn{2}{c}{\textit{Modality}} & \multicolumn{1}{c}{\multirow{2}{*}{\textit{Methods}}} \\
        \cmidrule(lr){1-3} \cmidrule(lr){5-6}
        \textit{Rule} &\textit{SFT} &\textit{RL} & & \textit{In} &\textit{Out} & \\ 
        \midrule

        \blueyes & -  & -  & & T  &  T  & Simul-LLM~\cite{agostinelli2024simul,raffel2024simultaneous}, StreamingGPE~\cite{tong2025llm}, STACL~\cite{ma2019stacl}, AsyncReasoning~\cite{yakushev2025asynchronous}, StreamingThinker~\cite{tong2025streamingthinker}, Conveyor~\cite{xu2024conveyor}, AsyncLM~\cite{gim2024asynchronous} \\
        \blueyes & -  & -  & & T  &  S  & CosyVoice 2~\cite{du2024cosyvoice}, IST-LM~\cite{yang2024interleaved}, DSM~\cite{zeghidour2025streaming} \\
        \blueyes & -  & -  & & S  &  T  & MFLA~\cite{xia2025mflamonotonicfinitelookahead}, InfiniSST~\cite{ouyang2025infinisstsimultaneoustranslationunbounded}, LLM as Processor~\cite{tong2025llm}, SASST~\cite{yang2025sasst}, SimulS2S-LLM~\cite{deng2025simuls2s}, ReaLLM~\cite{seide2024speech}, Llama-omni~\cite{fang2024llama} \\
        \blueyes & -  & -  & & S  &  S  & StreamRAG~\cite{arora2025stream} \\
        \blueyes & -  & -  & & V  &  T  & LiveCC~\cite{chen2025livecc},  StreamVLN~\cite{wei2025streamvln}, ActiveVLN~\cite{zhang2025activevln}, AViLA~\cite{zhang2025avila} \\
        
        - & \blueyes  & -  & & T  &  T  & SiLLM~\cite{guo2024sillm}, TransLLaMa~\cite{koshkin2024transllama}, EAST~\cite{fu2025llms}, DrFrattn~\cite{zhao2025drfrattn}, FineHarm~\cite{li2025judgment}, PsFuture~\cite{zhao2024psfuture}\\
        - & \blueyes  & -  & & T  &  S  & SimulMEGA~\cite{le2025simulmegamoeroutersadvanced}, Cosyvoice~\cite{du2024cosyvoice}, DSM~\cite{zeghidour2025streaming} \\
        - & \blueyes  & -  & & S  &  T  & Divergence~\cite{chen2024divergence}, SimulMEGA~\cite{le2025simulmegamoeroutersadvanced}, ReaLLM~\cite{seide2024speech}, Llama-omni~\cite{fang2024llama} \\
        - & \blueyes  & -  & & S  &  S  & StreamSpeech~\cite{zhang2024streamspeechsimultaneousspeechtospeechtranslation}, EASiST~\cite{fu2025efficient}, SimulMEGA~\cite{le2025simulmegamoeroutersadvanced}\\
    - & \blueyes  & -  & & V  &  T  & Videollm-online~\cite{chen2024videollm}, ProVideLLM~\cite{chatterjee2025streaming}, EyesWO~\cite{zhang2025eyes}, Streamo~\cite{xia2025streaming}, ProASIST~\cite{zhang2025proactive}, Videollm-MOD~\cite{wu2024videollm}, DisPider~\cite{qian2025dispider}, Stream-VLM~\cite{panchal2024say}, Lion-FS~\cite{li2025lion}, ProVideLLM~\cite{chatterjee2025streaming}, StreamBridge~\cite{wang2025streambridge}\\

        - & -  & \blueyes  & & T  &  T  & SeqPO-SiMT~\cite{xu2025seqpo}, Interleaved Reasoning~\cite{xie2025interleaved} \\
        - & -  & \blueyes  & & T  &  S  & Seed LiveInterpret 2.0~\cite{cheng2025seed} \\
        - & -  & \blueyes  & & S  &  T  & Seed LiveInterpret 2.0~\cite{cheng2025seed} \\
        - & -  & \blueyes  & & S  &  S  & Seed LiveInterpret 2.0~\cite{cheng2025seed} \\
        - & -  & \blueyes  & & V  &  T  & MMDuet2~\cite{wang2025mmduet2} \\

        \bottomrule
    \end{tabular}
}
\caption{Summary of additional literature on concurrent-streaming LLMs, complementing the discussion in Sec.~\ref{Sec:Concurrent-streaming}.}
\label{tab:appendix-B-con}
\end{table*}

\end{document}